\documentclass[10pt,journal,compsoc]{IEEEtran} 
\usepackage[T1]{fontenc}
\usepackage[utf8]{inputenc} 
\usepackage{authblk}

\usepackage{geometry} 
\usepackage{graphicx} 
\usepackage{multicol}

\usepackage{booktabs} 
\usepackage{array} 
\usepackage{paralist} 
\usepackage{verbatim} 
\usepackage{natbib}
\usepackage{hyperref}
\usepackage{tikz}

\usepackage{fancyhdr} 
\usepackage[nottoc,notlof,notlot]{tocbibind} 
\usepackage[titles,subfigure]{tocloft} 

\usepackage{amsmath}
\usepackage{amssymb}
\usepackage{amsthm}
\usepackage{mathrsfs}
\usepackage{enumerate}
\usepackage[all]{xy}
\usepackage{todonotes}
\usetikzlibrary{calc}
\usepackage{subcaption}

\usepackage[linesnumbered,ruled]{algorithm2e}

\theoremstyle{definition}

\newcommand{\wt}[1]{\widetilde{#1}}

\newcommand{\bbr}{\mathbb R}

\newcommand{\diff}{\backslash}

\newcommand{\sheaf}[1]{\mathcal{#1}}

\newcommand{\mxxx}[9]{\left(\begin{array}{ccc} #1 & #2 & #3 \\ #4 & #5 & #6 \\ #7 & #8 & #9 \end{array}\right)}

\newcommand{\nino}{\text{Ni\~{n}o }}
\newcommand{\nina}{\text{Ni\~{n}a }}

\begin{document}
\title{A New Spectral Clustering Algorithm}

\author{W.R.~Casper$^1$ and Balu~Nadiga{$^2$}
\thanks{$^1$Department of Mathematics, Louisiana State University, Baton Rouge LA. Email:wcasper1@lsu.edu}
\thanks{$^2$CCS-2, Los Alamos National Lab, Los Alamos, New Mexico. Email:balu@lanl.gov}
}


\IEEEtitleabstractindextext{
\begin{abstract}
  We present a new clustering algorithm that is based on searching for
  natural gaps in the components of the lowest energy eigenvectors of
  the Laplacian of a graph.  In comparing the performance of the
  proposed method with a set of other popular methods (KMEANS,
  spectral-KMEANS, and an agglomerative method) in the
  context of the Lancichinetti-Fortunato-Radicchi (LFR) Benchmark for
  undirected weighted overlapping networks, we find that the new
  method outperforms the other spectral methods considered in certain
  parameter regimes.  Finally, in an application to climate data
  involving one of the most important modes of interannual climate
  variability, the El \nino Southern Oscillation phenomenon, we
  demonstrate the ability of the new algorithm to readily identify 
  different flavors of the phenomenon.
\end{abstract}
}

\maketitle
\IEEEdisplaynontitleabstractindextext
\IEEEpeerreviewmaketitle

\ifCLASSOPTIONcompsoc
\IEEEraisesectionheading{\section{Introduction}\label{sec:introduction}}
\else
\section{Introduction}
\label{sec:introduction}
\fi
\IEEEPARstart{C}{lustering} is an unsupervised learning
technique used to identify natural subgroups of a set of data wherein
the members of a subgroup share certain properties.  On representing
the set of data as a graph, clustering methods group vertices of the
graph into clusters based on the edge structure of the graph
\citep[e.g., see][]{schaeffer2007graph}.  We note here that the edge
structure itself may be derived purely from interactions between
vertices\footnote{and in which context clustering is commonly referred
  to as community-detection}, e.g., as in a social network, or based
on a measure of similarity between the vertices themselves, or some
combination of the two. Needless to mention, clustering methods find
use in an extremely wide range of disciplines (e.g., neuroscience
\cite{galbraith2010study}, social networks
\cite{mishra2007clustering}, computer vision \cite{haralick1985image},
and digital forensic analysis \cite{decherchi2009text}).  Indeed most
modes of climate variability are typically first identified by
applying clustering methods to climatological data sets
\citep[e.g.,][and
others]{lund2009revisiting,corte1998regional,unal2003redefining,johnson2013many,
  steinhaeuser2011comparing}.

Since approaches to clustering are extremely varied, we do not attempt
even a brief overview of such methods. For that the reader is referred
to reviews such as \cite{schaeffer2007graph}. Instead, we shift focus
directly to the new algorithm after briefly considering common aspects
of spectral approaches. In this article, we propose a spectral
clustering algorithm based on direct analysis of the magnitudes of
components of the eigenvectors (of the Laplacian matrix; see next
section) with smallest eigenvalues.  We call this algorithm spectral
maximum gap method (SP-MGM).  We compare the performance of this
algorithm with popular clustering algorithms including k-means
(KMEAN), more typical spectral clustering based on running kmeans on
eigenvectors (SP-KMEANS), and agglomerative clustering (AGG).  To
determine the performance of these algorithms, we rely on a series of
LFR benchmark graphs \cite{lancichinetti2008benchmark}.  Notably, it
is seen that on a certain series of these graphs with varying size,
average degree, and weight mixing parameter, our algorithm
consistently outperform these other methods.  Following our benchmark
analysis, we apply SP-MGM to a collection of monthly averaged sea
surface temperatures in the \nino 3.4 region of the Pacific Ocean.  We
demonstrate that our method correctly and readily identifies the
various flavors of El \nino and La \nina events.

\section{Spectral Clustering Algorithms} Spectral clustering
algorithms rely on using spectral characteristics of the Laplacian of
a (weighted) graph to partition the vertices of the graph into natural
clusters.  Here, the Laplacian can refer to the normalized Laplacian,
the non-normalized Laplacian, or even stranger entities like the
$p$-Laplacian \cite{buhler2009spectral}.  The use of spectral
characteristics to identify clusters in a graph may be justified in
numerous ways, such as by considering random walks, minimal cuts, or
block matrix diagonalization.  For a recent survey on spectral
clustering methods, the reader is referred to
\cite{nascimento2011spectral}; for an introduction to spectral
clustering itself, see, e.g., \cite{von2007tutorial}.

To present a brief and intuitive description of spectral clustering
methods, we consider the eigenvectors of the Laplacian of a
disconnected graph.  Let $G$ be a graph with adjacence matrix $A$ and
degree matrix $D$ (the diagonal matrix whose nonzero entries are the
degrees of the vertices of $G$).  Then the (non-normalized) Laplacian
of $G$ is defined as $L=D-A$.  The matrix $L$ is positive
semidefinite, and the kernel has a basis spanned by the indicator
functions of the connected components of $G$.  In particular, if $G$
is disconnected then the connected components of $G$ may be deduced
from the kernel of $L$.

Clusters in a graph $G$ are intuitively regions in $G$ where the vertices are
connected by edges with strong weights, relative to the weights of edges going
between clusters.  For this reason, the Laplacian $L$ of the graph $G$ should
be very close to the Laplacian $\wt L$ of the disconnected graph $\wt G$ formed
from $G$ by deleting intercluster edges.  This means that the eigenvalues of $L$
will correspond approximately to the eigenvalues of $\wt L$.  Since $\wt L$ is
positive semidefinite, the smallest eigenvalue of $\wt L$ is zero.  Therefore
the eigenvectors of the smallest eigenvalues of $L$ should be small
perturbations of vectors in the kernel of $\wt L$.  Thus by analyzing the
structure of the eigenvectors corresponding to the smallest handful of
eigenvalues of $L$, we should expect to retrieve information about the clusters
of $G$.  It is exactly this idea that all spectral clustering algorithms
exploit.  Note that if $G$ is connected, then the kernel of $L$ consists of
vectors whose entries are all identical, and therefore does not provide any
information.  For this reason, the kernel of $L$ is typically ignored in
spectral clustering algorithms.

\begin{figure}[htp]
\begin{center}
\begin{subfigure}{0.4\textwidth}
\includegraphics[width=\linewidth]{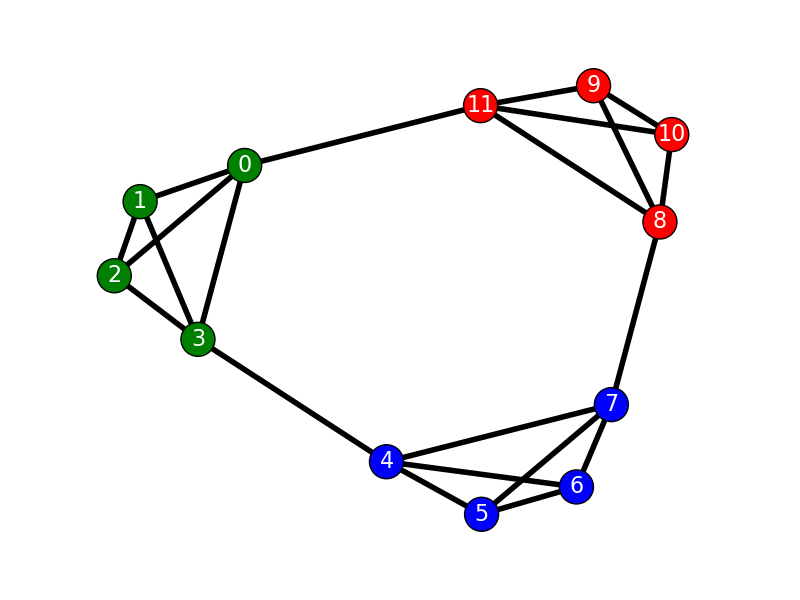}
\caption{A graph with three clusters formed by cliques of size $4$.  Vertex
color indicates inclusion in a particular clique, and all edges have weight
$1$.  The nodes are also indexed in the same order as their appearance in the
Laplacian matrix.}\label{cliquegraph}
\end{subfigure}
\begin{subfigure}{0.4\textwidth}
\includegraphics[width=\linewidth]{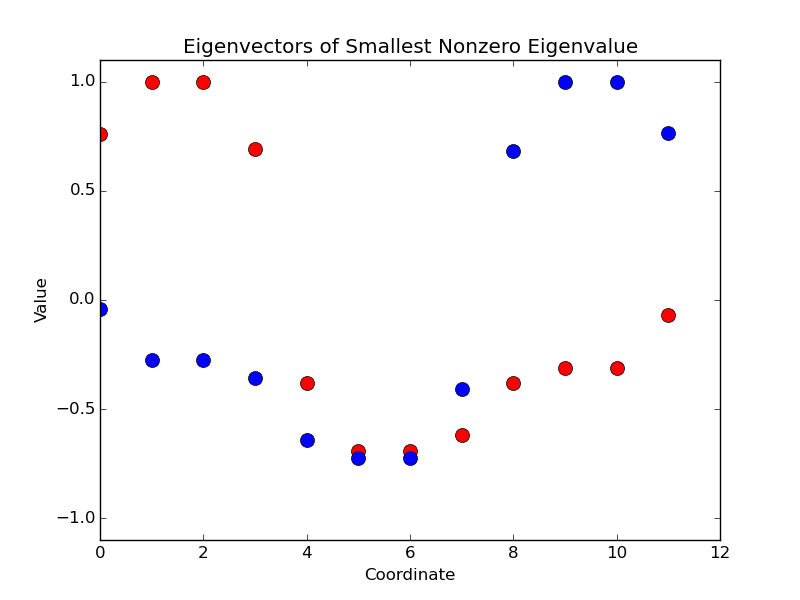}
\caption{A basis for the eigenspace of eigenvectors of the clique
  cluster graph with smallest nonzero eigenvalue $\lambda \approx 0.5051$.  The
  eigenspace is two-dimensional, and the basis elements are color-coded.  The
  structure of these eigenvectors reflects cluster membership in that the
  values (for any given eigenvector) at indices belonging to the same cluster
  are similar.  }\label{cliquegraphevecs}
\end{subfigure}
\caption{An illustration of the basis for spectral clustering.}
\end{center}
\end{figure}

To further clarify this idea, consider the example of three cliques of size $4$
joined together by single bonds in general position, as in Figure
\ref{cliquegraph}.
The non-normalized Laplacian of this graph in block matrix form is
$$L = \mxxx{K + K'}{Q}{Q^T}{Q^T}{K + K'}{Q}{Q}{Q^T}{K+K'},$$
$K$ the Laplacian of a $4\times 4$ clique, $K' = QQ^T+Q^TQ$, and $Q$ the
$4\times 4$ matrix whose only nonzero entry is $-1$ and occurs in the bottom
left corner.  The eigenvalues of this matrix are $0$ and $6$ with multiplicity
$1$, $6-\lambda$ and $\lambda$ with multiplicity $2$, and $4$ with multiplicity
$6$, where here $\lambda \approx 0.5051$.  A basis for the for the eigenspace
of the lowest nonzero eigenvalue $\lambda$ is plotted in Figure
\ref{cliquegraphevecs}.  The stucture of these eigenvectors successfully
reveals the presence of the three clusters.

In our algorithm, spectral clustering on a
given data set occurs in three parts.  We first create a similarity
matrix $A$ describing the similarity or connectedness between each of
our pieces of data.  Secondly, we calculate the eigenvectors of the
Laplacian of $A$, and sort them in terms of the magnitudes of the
associated eigenvalues.  Lastly, we use some number of the
eigenvectors with the smallest nonzero eigenvalues to calculate a
natural partition of the vertices of the graph.

A similarity matrix for a collection of $n$ indexed data points is an
$n\times n$ symmetric matrix $A$ whose $i,j$'th entry is a nonnegative
value measuring the similarity between the $i$ and $j$'th data values.
It is assumed that larger values indicate more similar objects, and
the diagonal elements are typically taken to be $0$.  Similarity
matrices can be created out of data in a variety of ways.  One popular
method is to start with the distance $d_{ij}$ between $i$ and $j$
(under some metric), and then define $A_{ij} =
\exp(-d_{ij}^2/2\sigma)$ for $i\neq j$, where here $\sigma$ is a
parameter.

The (symmetric) normalized Laplacian of $A$ is the graph Laplacian of
the unique weighted graph $G(A)$ whose weight matrix is $A$.  In
particular, it is defined by
$$L = I - D^{-1/2}AD^{-1/2}$$
where here $D$ is a diagonal matrix whose entries are the sums of each
of the column vectors of $A$.  The matrix $L$ is positive-semidefinite
and will always have at least one eigenvector $\vec v$ with eigenvalue
$0$, given by $\vec v = D^{1/2}[1\ \dots\ 1]^T$.  However, assuming
that the graph $G(A)$ associated to $A$ is connected, this will be the
only eigenvector with eigenvalue $0$, up to constant multiples.

The smallest nonzero eigenvectors of $L$ encode clustering information
for $L$, and it is these eigenvectors which we wish to parse in the
final step.  The reason these eigenvectors contain clustering
information can be made intuitive by the following argument.  Clusters
consist of regions in the graph with strong interconnectivity.
Specifically, given a vertex in a cluster we would expect that on
average the strength of its connection with other vertices in the same
cluster should be much greater than the strength of its connection
with vertices in other clusters.  From this point of view, we can
decompose the adjacency matrix $A$ as $A = B+C$ where $B$ is a block
diagonal matrix whose blocks are formed by eliminating inter-cluster
bonds, and where $C=B-A$ is relatively small compared to $B$.  In this
case, the Laplacian $L$ of $A$ may be written as $L=L_B + L'$ where
$L_B$ is the Laplacian of $B$ and $L'$ is small relative to $L_B$.
Consequently the eigenvectors of $L_B$ will be comparable ot the
eigenvectors of $L$.  In particular the $0$ eigenvectors of $L_B$
correspond precisely with the eigenvectors of $L$ with small
eigenvalues.  Moreover, these eigenvectors describe the connected
components of the graph $G(B)$ of $B$, and these are precisely the
clusters of $G(A)$.  In this way, using the eigenvectors of the
smallest nonzero eigenvalues makes sense.

Where various clustering algorithms differ is in this third step,
where the eigenvector data is used to calculate clusters.  Before proposing our
own clustering scheme, we will recount two common methods found in the
literature, the simplest method SP-G1 and the most common method SP-KMEANS.

\subsection{SP-G1 clustering}
The oldest and most basic algorithm is SP-G1, which decomposes the data
into two clusters based on the entries of the eigenvector
$\vec v = [v_1\ \dots\ v_n]^T$ with the lowest nonzero eigenvalue
\cite{fiedler1975property}\cite{jain1988algorithms}.  Specifically, a
theshold value $r$ needs to be chosen.  Then the $i$'th data point is put into
the first cluster if $v_i < r$, and otherwise it is put into the
second cluster.  Of course, there are several natural choices for $r$,
including $r = n^{-1}\sum_{i}v_i$.  Of course, more sophisticated
methods for the choice of $r$ exist\cite{shi2000normalized}.

\begin{algorithm}
    \caption{SP-G1 clustering} \SetKwInOut{Input}{Input}
\SetKwInOut{Output}{Output} \Input{weight matrix $A$} \Output{two
clusters} Calculate normalized Laplacian $L$ of $A$\\ Calculate
smallest positive eigenvalue $\lambda_1$ of $L$\\ Calculate an
eigenvector $\vec v$ of $\lambda_1$\\ Choose a threshold $r$\\
\eIf{$v_i<r$} { put node i in cluster 1 } { put node i in cluster 2 }
\end{algorithm}

\subsection{SP-KMEANS}
A more complicated and very reliable clustering algorithm instead uses
eigenvectors $\vec v_1,\dots,\vec v_k$ of the $k$ smallest positive
eigenvalues $0<\lambda_1\leq\dots\leq \lambda_k$ of the Laplacian $L$
of $A$.  These eigenvectors are used as the column vectors of an
$n\times k$ matrix $V$.  The matrix $V$ is then normalized so that
each of the rows has norm $1$, and the associated row vectors are
viewed as $n$ points on the surface of a sphere in $\bbr^k$.  Then by
running KMEANS on these points, we return $k$ clusters of the data
$C_1,\dots, C_k$ which partition the original graph
\cite{ng2002spectral}\cite{meila2001learning}.  The reliability and
familiarity of this algorithm has lead to its popularity as one of the
most common forms of spectral clustering, with implementations in both
R and python \cite{scikit-learn}\cite{kernlab}.

\begin{algorithm}
    \caption{SP-KMEANS clustering} \SetKwInOut{Input}{Input}
\SetKwInOut{Output}{Output} \Input{weight matrix $A$\\ number of
clusters $k$} \Output{$k$ clusters} Calculate normalized Laplacian $L$
of $A$\\ Calculate $k$ smallest positive eigenvalues
$\lambda_1<\dots<\lambda_k$ of $L$\\ Calculate an eigenvector $\vec
v_j$ of $\lambda_i$\\ Construct an $n\times k$ matrix $V= [\vec v_1\
\dots\ \vec v_k]$\\ Normalize $V$ so that each row vector has norm
$1$\\ Run KMEANS on the row vectors of $V$
\end{algorithm}

\subsection{Spectral Maximum Gap Method (SP-MGM)} 
In the new SP-MGM algorithm we propose, the eigenvectors
$\vec v_1,\dots,\vec v_k$ of the $k$ smallest positive eigenvalues
$0<\lambda_1\leq\dots\leq \lambda_k$ of the Laplacian $L$ of $A$ are
first calculated.  Next, for each $j$ we create a new vector
$\vec u_j$ whose entries are the entries of $\vec v_j$ in increasing
order.  Let $v_{ji}$ and $u_{ji}$ denote the entries of $\vec v_j$ and
$\vec u_j$, respectively.  Differences $u_{j(i+1)}-u_{ji}$ between
subsequent entries of $\vec u_j$ represent ``gaps" in $\vec v_j$.

We choose $\ell_j$ so that $\vec u_{j(\ell_j+1)}-\vec u_{j\ell_j}$ is
the largest gap.  Then we create a partition $P_j,Q_j$ of $\{0,\dots,
n-1\}$ by placing vertex $i$ in $P_j$ if $v_{ji} <= u_{j\ell_j}$, and
placing vertex $i$ in $Q_j$ otherwise.  Finally, for each subset $S$
of $U=\{1,\dots,k\}$ we define a cluster
$$C_S = \left(\bigcap_{i\in S}P_i\right)\cap\left(\bigcap_{i\in U\diff S}Q_i\right),$$
where here if $S$ or $U\diff S$ is empty, the associated intersection
is interpreted to be $\{0,1,\dots,n-1\}$.  It is clear from the
construction that $\{C_S: S\subseteq U\}$ forms a partition of
$\{0,1,\dots,n-1\}$.

The number of clusters in this partition is at least $2$, but can be
up to $2^n$.  However, in practice it is observed that many of the
clusters $C_S$ are repeated, and the algorithm tends to find far fewer
than $2^n$ clusters.  In fact, the number of clusters created in this
way tends to be close to the actual right number of clusters.  This
will be discussed further in the benchmark section below.  Note that
we experimented with both using the normalized and non-normalized
Laplacian.  We found that the non-normalized Laplacian performs better
for the benchmark graphs that we considered.  We note that a common and
effective strategy in applying spectral methods to large data sets
consists of adopting a hierarchical approach wherein spectral
techniques are used on subsets of the data that are then recombined
\citep[e.g., see][]{luxburg2005limits, tung2010enabling, semertzidis2015large, peluffo2014short}.

\begin{algorithm}
    \caption{SP-MGM clustering} \SetKwInOut{Input}{Input}
\SetKwInOut{Output}{Output} \Input{weight matrix $A$\\ number of modes
$k$} \Output{$m$ clusters ($2\leq m\leq 2^k$)} Calculate
non-normalized Laplacian $L$ of $A$\\ Calculate $k$ smallest positive
eigenvalues $\lambda_1<\dots<\lambda_k$ of $L$\\ Calculate an
eigenvector $\vec v_j$ of $\lambda_i$\\ Set $\vec u_j =
\text{sort}(\vec v_j)$\\ Choose $\ell_j$ so that
$u_{j(\ell_j+1)}-u_{j\ell_j}$ is largest\\ \eIf{$v_{ji}<u_{j\ell_j}$}
{ put node i in cluster $P_j$ } { put node i in cluster $Q_j$ } For
each $S\subseteq \{1,\dots,k\}$ define
    $$C_S = \left(\bigcap_{i\in S}P_i\right)\cap\left(\bigcap_{i\in U\diff S}Q_i\right),$$
\end{algorithm}

\subsubsection{A mathematical justification of the new algorithm}
We provide a mathematical justification of the SP-MGM algorithm by applying the
theory of eigenvector permutations to the Laplacian of the graph $G$.  We
provide full mathematical details in the appendix for the interested
reader. However, we restrict ourselves to a brief description of the
justification here so as to keep the narrative more generally
accessible: First, we show mathematically that if $G$ consists of
several well-defined clusters $C_1,\dots, C_k$ of size $n_1,\dots,n_k$
then 
$$\max_{i\neq m} (n_i-1)\frac{\rho_1}{\lambda_{i,2}} <
\frac{1}{4n_m^{1/2}},$$ for all $1\leq m\leq k$. 
Here $\lambda_{i,2}$ and $\rho_1$ are spectral data which
measure the quality of the clusters.  Specifically, $\lambda_{i,2}$ is
the smallest nonzero eigenvalue of cluster $C_i$ and measures the
connectivity of the cluster $C_i$.  For this reason, it is sometimes
called the algebraic connectivity of the cluster $C_i$.  The number
$\rho_1$ is the spectral radius of the Laplacian of the graph $G$
formed by deleting all of the intra-cluster edges in $G$.  In this
way, it measures the strength of the inter-cluster connections.  For a
strongly clustered graph, it therefore makes sense that $\rho_1$ is
small relative to $\lambda_{i,2}$ for all $i$.  Further, if $G$
consists of several {\em such} clusters $C_1,\dots, C_k$, then, we can
mathematically {\em guarantee} that SP-MGM will correctly identify the
clusters.  Again, the reader is
referred to the appendix for details.

\section{Benchmark Experiments}
\subsection{LFR Benchmark} To test the quality of our
clustering/community-detection algorithm, we choose to use the LFR
benchmark.  Benchmark graphs consist of randomly generated graphs with
built-in network ties.  The properties of the output graph are
controlled by a series of parameters. The parameters we consider and
vary are the total number of nodes $n$, the average degree $d_{avg}$,
the maximum degree $d_{max}$, and the mixing parameter $\mu$.  Each
node is given a degree based on a power law distribution.  The minimum
and maximum degrees in the network are chosen so that the mean from
the power law distribution is $d_{avg}$.  Nodes are assigned to
clusters such that the (weighted) fraction of intercluster edges is
$0\leq \mu \leq 1$.  Source code for generating LFR Benchmark graphs
is available at
\url{https://github.com/eXascaleInfolab/LFR-Benchmark_UndirWeightOvp}.

Note also that each of the algorithms described above requires an important piece of user input: namely the number of clusters to look for.  In practice, there are several ways of deciding how many clusters in a graph to look for, such as the elbow technique, the jump method, and silhouette analysis \cite{kodinariya2013review}.  However, these seem outside the scope of the current paper.  Instead, we supply each algorithm with the correct number of clusters to look for.  The exception to this is SP-MGM, which does not take in a number of clusters to look for.  Instead, it takes in a number of eigenmodes $m$.  However, in practice, the number of clusters it finds is $m+1$, so we choose $m$ accordingly.

\subsection{The Quality of a Clustering Partition} After a clustering
algorithm identifies one or more clusters in a graph, it is natural to
want to identify the quality of the clusters.  Are the vertices in a
cluster naturally correlated in some fashion, or is the cluster only
an artifact of the specified clustering algorithm?  While there is no
single criteria for reliably determining the quality of a cluster,
there exist several metrics which may be used in combination to judge
the reasonability of ones choice of clusters.  For a comparison of
several such metrics, see \cite{almeida2011there}.

The LFR Benchmark produces a weighted graph,
along with a predetermined true value for the neighborhoods.  To
determine the performance of each clustering algorithm we use two
scoring metrics: the normalized mutual info score (NMI) and the
adjusted rand score (ARS).  In both metrics, the scores can range
between $0$ and $1$, with $1$ representing perfect cluster prediction
and $0$ representing largely independent cluster prediction from the
true clusters.

\subsection{Results}
We will now compare the performance of our algorithm to the
performance of a traditional spectral clustering SP-KMEANS, as well as
traditional KMEANS and an agglomerative clustering algorithm.  This
collection of algorithms was chosen as a basis of comparison since
they are notably diverse in their methodologies. At the same time,
they are also popular enough to have been implemented in the widely-used
python machine learning package scikit-learn
\cite{scikit-learn}.  While SP-KMEANS has already been described, the reader is
referred to \cite{rokach2005data} for descriptions of 
KMEANS and the agglomerative methods.

\subsubsection{Variations in Mixing Parameter}
\begin{figure}[htp]
\includegraphics[width=\linewidth]{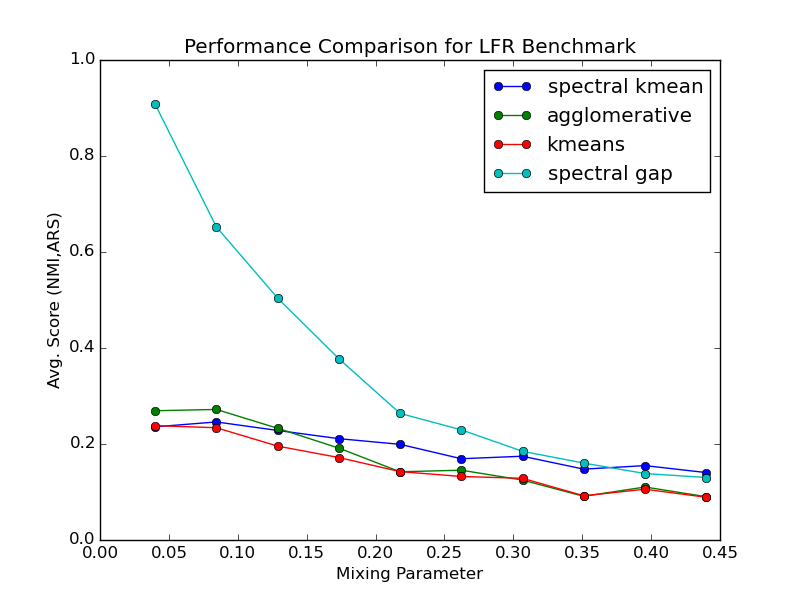}
\caption{The average of the normalized mutual info score (NMI) and
adjusted rand score (ARS) for various values of the mixing parameter
$\mu$.  The benchmark graphs used here were all $50$ nodes each with
an average degree of $5$ and a maximum degree of $20$.  The scores for
each value of $\mu$ are averaged over $50$ randomly generated
benchmark graphs.}\label{benchmark50}
\end{figure}

\begin{figure}[htp]
\includegraphics[width=\linewidth]{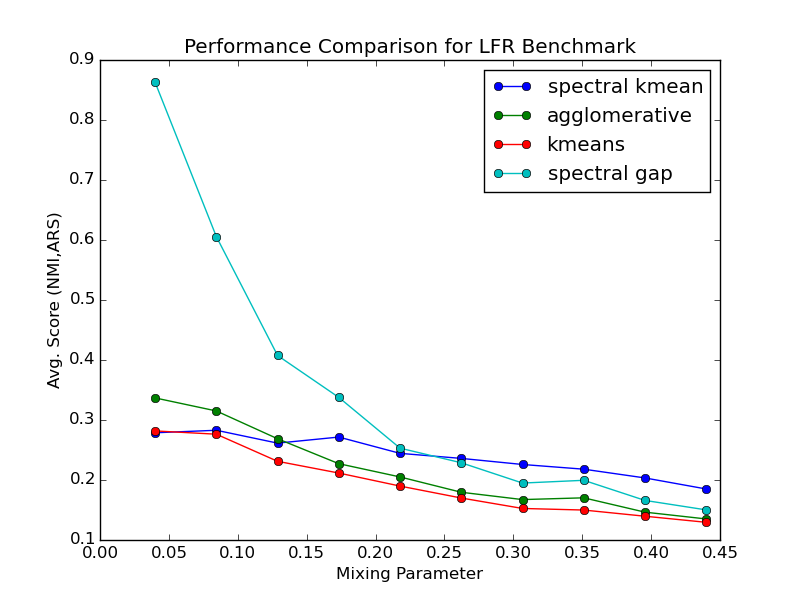}
\caption{The average of the normalized mutual info score (NMI) and
adjusted rand score (ARS) for various values of the mixing parameter
$\mu$.  The benchmark graphs used here were all $100$ nodes each with
an average degree of $5$ and a maximum degree of $20$.  The scores for
each value of $\mu$ are averaged over $50$ randomly generated
benchmark graphs.}\label{benchmark100}
\end{figure}

The scores of clustering algorithms for the LFR benchmark on graphs
with $50$ and $100$ nodes with varying mixing parameter are shown in
Figures \ref{benchmark50} and \ref{benchmark100}.  From these figures,
it is evident, and notably so, that SP-MGM consistently outperforms
the other clustering algorithms considered for small networks over the
entire range of the mixing parameter considered.

\begin{figure}[htp]
\includegraphics[width=\linewidth]{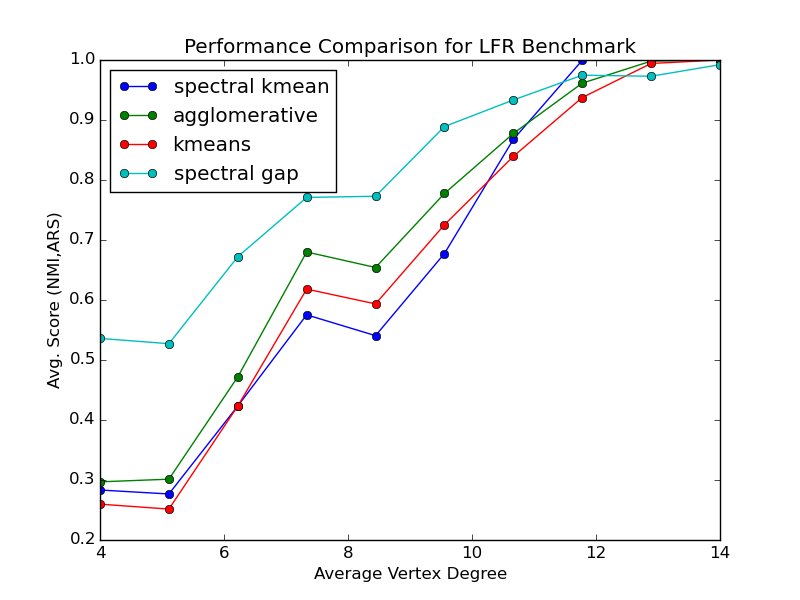}
\caption{The average of the normalized mutual info score (NMI) and
adjusted rand score (ARS) vs. average vertex degree $k$.  The
benchmark graphs used here all had mixing parameter $\mu=0.1$ and
network size $n=100$.  Each score is averaged over $50$ randomly
generated benchmark graphs.}\label{benchmark2_100}
\end{figure}
\begin{figure}[htp]
\includegraphics[width=\linewidth]{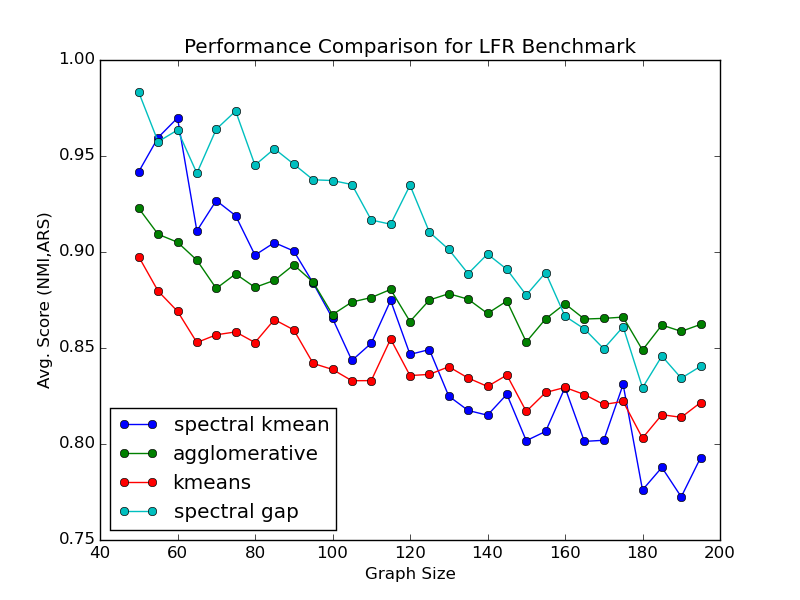}
\caption{The average of the normalized mutual info score (NMI) and
adjusted rand score (ARS) for various network sizes $n$.  The
benchmark graphs used here all had mixing parameter $\mu=0.1$ and
average connectivity $k=10$.  Each score is averaged over $100$
randomly generated benchmark graphs.  The data shows that the spectral
gap method tends to have comparable or better performance than other
methods for a wide range of network sizes. However, performance worsens for
larger network sizes.}\label{benchmark3}
\end{figure} 

Next, we examined the performance of SP-MGM relative to the other
algorithms considered while varying the average vertex degree, and the
network size.  Figure \ref{benchmark2_100} shows the results when
average vertex degree was varied. As the average network degree
increased, all algorithms considered showed an increase in score.
However, SP-MGM consistently out-performed the other algorithms
considered.  Furthermore, SP-MGM was the only algorithm to perform
reasonably when the average vertex degree was small, and the mixing
parameter was simultaneously low.

The scores of clustering algorithms for the LFR benchmark on graphs
with $50$ and $100$ nodes with varying mixing number are shown in
Figures \ref{benchmark50} and \ref{benchmark100}.  From these figures,
it is evident, and notably so, that SP-MGM consistently outperforms
the other clustering algorithms considered for small networks over the
entire range of the mixing parameter considered.

For larger networks (not shown), SP-MGM performs comparably or better
than the other algorithms considered when the algorithms are scoring
high enough.  In other words, when any of the algorithms perform well,
SP-MGM performs well as well. We note here that when edges are
distributed uniformly across the nodes, clusters are not well-defined,
and as such, the computed clustering is likely not robust (and likely
arbitrary) and therefore not meaningful or useful. Notably, for small
values of the mixing parameter on small graphs, the prediction-skill
of SP-MGM is far superior.

\subsubsection{Variations in Average Degree and Network Size}

Next, we examined the performance of SP-MGM relative to the other
algorithms considered while varying the average vertex degree, and the
network size.  Figure \ref{benchmark2_100} shows the results when
average vertex degree was varied. As the average network degree
increased, all algorithms considered showed an increase in score.
However, SP-MGM consistently out-performed the other algorithms
considered.  Furthermore, SP-MGM was the only algorithm to perform
reasonably when the average vertex degree was small, and the mixing
parameter was simultaneously low.

Figure \ref{benchmark3} shows the results when the network size was
varied. SP-MGM is seen to perform well over a range of network sizes,
with only the agglomerative clustering algorithm out-performing it for
larger graph sizes.  We note, however, that the new algorithm tended
to outperform its spectral counterpart SP-KMEAN over the full range of
network sizes considered.


\subsubsection{The Number of Clusters SP-MGM Detects}
From the description of the SP-MGM algorithm, it is seen that we need
to specify only the number $m$ of eigenmodes to consider.  For this
reason, a priori the algorithm should be able to predict anywhere from
$2$ and $\min\{2^m,n\}$ clusters, where here $n$ is the total number
of vertices in the graph.  However, in practice the algorithm almost
always produces $m+1$ clusters, at least for all the test graphs
examined in this paper.  This result is quite surprising, since from
an intuitive perspective one would suspect the number of clusters
produced to be far more volatile.  It seems like this is additional
computational evidence that the ``maximum gaps" we are computing must
have a lot to do with the structure of the graphs themselves, closely
tied to the clusters.

\section{Clustering of SST data in El Ni\~{n}o 3.4 }
\begin{figure*}[htp]
\includegraphics[width=\linewidth,clip,trim={0 1cm 0 0}]{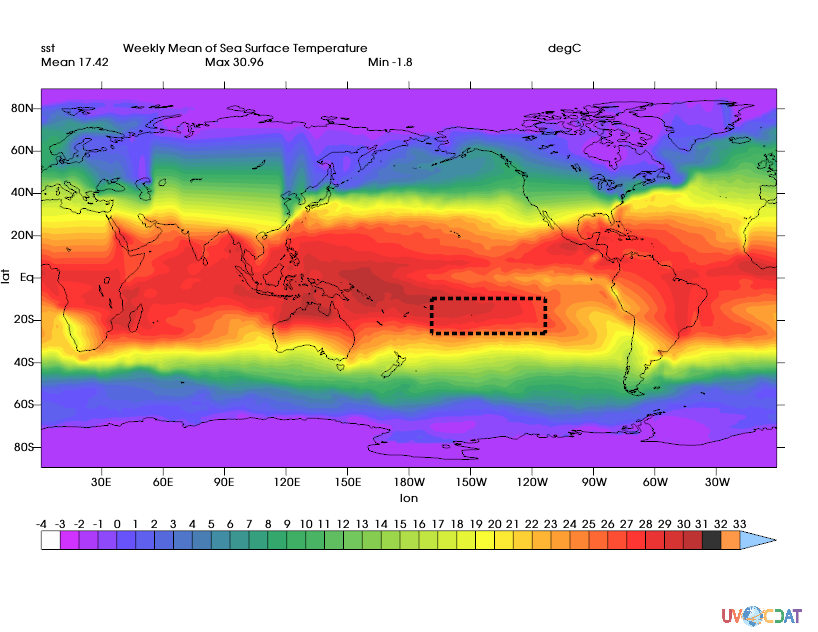}
\caption{Monthly averaged sea surface temperature on Jan 1, 2017.  The
region surrounded by the dashed line is the Ni\~{n}o 3.4 region, which
lies in 170W-120W, 5S-5N. }\label{sstplot}
\end{figure*}
In this section we consider the application of the new clustering
algorithm to a real world data set in the context of the climate
system.  Specifically we consider the El \nino Southern Oscillation
(ENSO) since it is one of the most important modes of interannual
variability of the climate system. It can be characterised in terms of
the sea surface temperature anomaly occuring in a region of the
equatorial Pacific Ocean.  The ENSO temperature anomaly roughly
fluctuates between three phases: a warm El \nino phase, a cold La
\nina phase, and a neutral phase.  The phases of ENSO are correlated
with changes in precipitation in various regions around the globe,
with strong correlations in coastal Pacific regions. As such,
predictions of ENSO have high socio-economic value, Luckily, ENSO also
happens to be one of the few modes of interannual variability that has
a useful level of verified predictability.

\subsection{Sea Surface Temperature Data}
We use version 4 of the Extended Reconstructed Sea Surface Temperature
(ERSST) dataset available at
\url{https://www.esrl.noaa.gov/psd/data/gridded/data.noaa.ersst.v4.html}.
The data has a spatial resolution of 2 degree, and while the
monthly-averaged data that we use are available from 1854-present, we
focus attention on data over the years 1955-2016, given lower
uncertainties over this more recent (instrumented) period. For further
description of the data, the reader is referred to
\url{https://www.ncdc.noaa.gov/data-access/marineocean-data/extended-reconstructed-sea-surface-temperature-ersst-v4}.

\subsection{El Ni\~{n}o Events}
\begin{figure*}[htp]
\includegraphics[width=\linewidth]{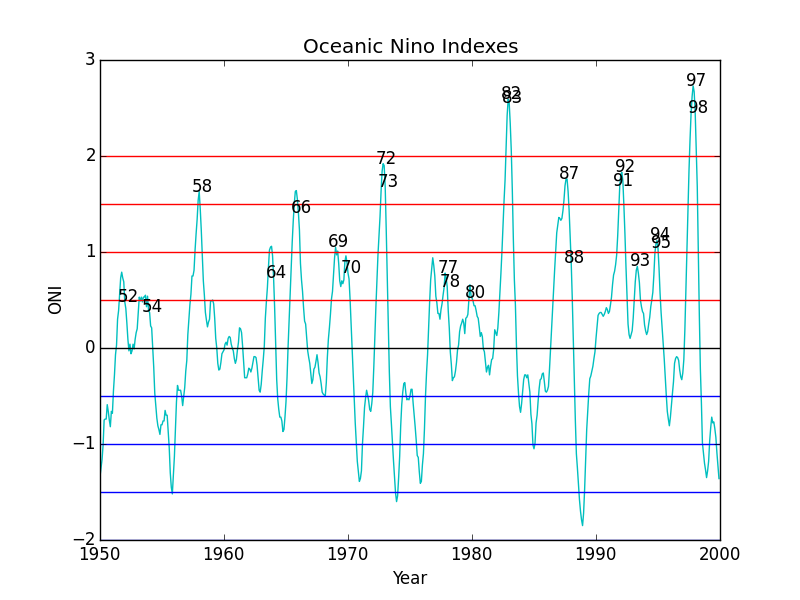}
\caption{Values of the Oceanic Ni\~{n}o Index (ONI) between 1950 and
2000.  Years where an El Ni\~{n}o event occurred are marked. Since
impacts of El \nino are largest in the (northern-hemisphere), winter
some of the events span the calendar year boundary.}
\label{nino3.4}
\end{figure*}

The average sea surface temperature is calculated using
a $30$-year average of the NOAA ERSST data starting Jan 1, 1950.  SST
anomaly is calulated relative to this mean
\cite{trenberth1997definition}.

We consider the data from ERSSTv4 in the Ni\~{n}o 3.4 region (the
region between 170W and 120W longitude and between -5N and 5N
latitude), as shown in Figure \ref{sstplot}.  El
Ni\~{n}o events are classified by a persistent large average
temperature anomaly in this region.  Quantitatively, El Ni\~{n}o
events are determined by the Oceanic Ni\~{n}o Index (ONI), which is a
$3$-month running mean of the SST anomaly spatially averaged 
over the Ni\~{n}o 3.4 region.  A Ni\~{n}o event is characterized by ONI
values greater than $0.5$ degrees C for at least three months in a
row.  The years of El Ni\~{n}o events between 1950 and 2000 may be
seen from the graph in Figure (\ref{nino3.4}) below.

\subsection{Clustering Results}
\begin{figure*}[htp]
\includegraphics[width=\linewidth]{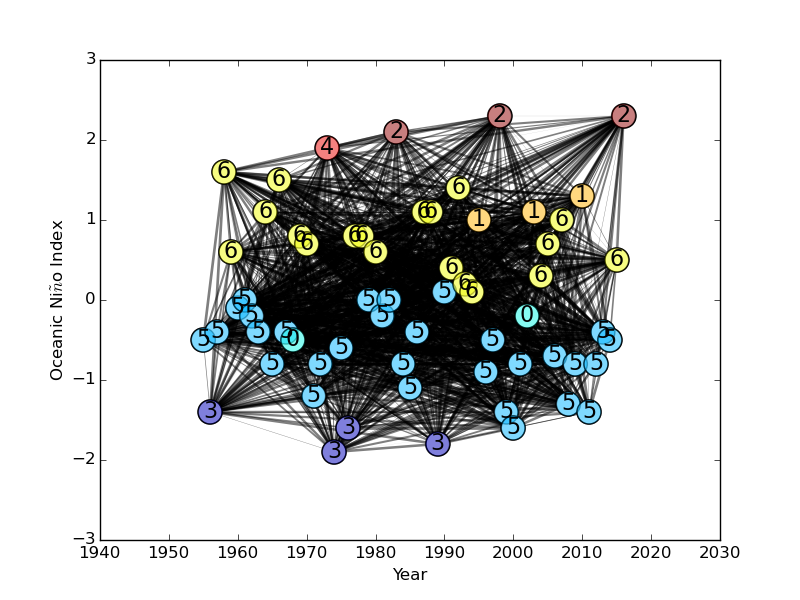}
\caption{Results of clustering NDJ SST states in the Ni\~{n}o 3.4
region.  Nodes are colored based on the cluster to which they belong.
Years with similar ONI tend to be clustered together.}\label{cluster}
\end{figure*}

While ONI is a conventional and simple measure that is used to classify the
state of ENSO, the phenomenon itself is more complicated. That is,
although the ENSO phenomenon can be characterized as a low-dimensional
dynamical phenomenon occuring in an infinite-dimensional dynamical
system, it is unlikely to be fully or adequately characterized by just
ONI. As such, we conduct a clustering analysis of the
spatially-extended sea surface temperature anomaly (SSTA) dataset T(x,
y, t) where t runs from years 1955 through 2016 and x and y correspond
to the \nino 3.4 region described above.  The data vector for each
year itself is a 3-month average starting from November of the
previous year.

In order to conduct cluster analysis, we next transform the SSTA
dataset into a graph whose vertices are the yearly SSTA datapoints
embedded in a high-dimensional space, the number of dimensions
corresponding to the number of distinct spatial locations in the \nino
3.4 region. Next we consider an edge structure that reflects the
mutual similarity of pairs of vertices. The resultant similarity
matrix has the form $A_{ij} = \exp(-d_{ij}/\sigma)$ for $i\neq j$,
where $d_{ij} = \vec T(x, y, i)^T\Sigma(x,y)^{-1}\vec T(x, y, j)$, and
where $\Sigma$ the spatially-extended covariance matrix for the period
considered. Note that $\Sigma(x,y)^{-1}$ is also sometimes called the
precision matrix. For simplicity, we consider a diagonal form for
$\Sigma(x,y)$ that comprises the variances of the SSTA at each of
the spatial locations in the \nino 3.4 region.  The parameter $\sigma$
toggles controls localization of the similarity matrix; we use
$\sigma=10^{-2}$.

In Figure \ref{cluster}, the nodes are colored-coded based on
cluster-membership and the nodes are placed at the two-dimensional
point corresponding to the year (x-axis) and ONI of that year
(y-axis). Finally, the thickness of the edges reflect similarity
between the pair of vertices they connect. In this figure, it is seen
that there is good correlation between the clusters and the
conventional ONI (y-axis).  Thus, the new clustering algorithm SP-MGM
is seen to successfully distinguish between El \nino, La \nina, and
neutral events.

Next, it is seen that the SP-MGM clustering algorithm identifies $7$
distinct clusters and that cluster membership is not solely determined
by ONI. Indeed, that there are different flavors of ENSO
besides El \nino and La \nina is now widely recognized and how these
different flavors influence climate in different regions is an
active area of research.  The cluster patterns we find are consistent,
e.g., with the patterns found in \cite{johnson2013many} who use a
neural-network based analysis in conjunction with other statistical
distinguishability tests. We note, however, that there are significant
differences in the data used by \cite{johnson2013many} in that they
consider a six month average starting in September and a much more
extensive region of the Pacific.

\section{Conclusions} 

In this paper, we presented a new spectral clustering algorithm which
we call the Spectral-Maximum Gap Method SP-MGM. This algorithm is
based on identifying gaps in the structure of eigenvectors.  We then
went on to provide a mathematical justification for the
algorithm. While it would be interesting to see where this algorithm
fits in a systematic comparison of spectral algorithms, such as in
\cite{verma2003comparison}, such a comparison is outside the scope of
the current paper. Further, as with other spectral schemes, this
method can be applied to large datasets in conjunction with a
hierarchical approach as in \citep[][and others]{luxburg2005limits,
  tung2010enabling, semertzidis2015large, peluffo2014short}.

Next, we examined the performance of this algorithm in comparison to a
few other popular algorithms using the LFR benchmark graphs.  Results
showed that the SP-MGM algorithm performs either comparbly or better
than its counterparts in a variety of parameter regimes.  We also
demonstrated how SP-MGM may be used to automatically detect an
appropriate number of clusters for a specified graph.  Finally, we
applied this algorithm to analyze and identify a variety of flavors of
the El \nino Southern Oscillation using spatially extended sea surface
temperature data.

\section*{Acknowledgement} This work was performed during one authors
visit to CNLS at Los Alamos National Lab.

\appendix
\subsection*{Eigenvector Perturbation}
The eigenvalue perturbation problem is the problem of estimating the
eigenvectors and eigenvalues of a perturbed matrix $T = T_0 + \epsilon
T_1$ based on the eigenvalues of the matrix $T_0$, where here
$\epsilon$ is a small positive number which estimates the size of the
perturbation.  In such a case one should rightfully assume that the
eigenvalues and eigenvectors of $T$ and $T_0$ are nearly the same.  In
this section, we briefly recount the idea of eigenvector perturbation
analysis, deriving in particular the linear perturbation equations for
the eigendata of a perturbed matrix.  This equation specifically
relates the eigenvectors of $T_0$ and $T = T_0+\epsilon T_1$ for
$\epsilon$ sufficiently small.  Our exposition is loosely based on
that found in the standard source \cite{demmel1997applied}.  However,
standard perturbation analysis formulas seems to emphasize the
reqirement that the eigenvalues of $T_0$ are distinct, whereas our
brief derivation shows that this is not the case, as long as the
formulas are appropriately adjusted.  This is important for our
application to graph theory, where we wish $T_0$ to be the Laplacian
of a disconnected graph.

Assume that $T_0,T_1$ are symmetric $n\times n$ matrices and let $\vec
u_1,\dots, \vec u_n$ be an orthonormal eigenbasis for $T_1$, with
$T_1\vec u_i = \lambda_i \vec u_i$.  Then the linear perturbation
equation says that for all $i$ there exists an eigenvector $\vec v_i$
of $T$ satisfying
\begin{equation}\label{linear eigenvector perturbation} \vec v_i =
\vec u_i + \sum_{j:\lambda_j\neq\lambda i}\frac{\epsilon\langle \vec
u_j,T_1\vec u_i\rangle}{\lambda_j-\lambda_i}\vec u_i + \sheaf
O(\epsilon^2).
\end{equation} Furthermore the eigenvalue $\lambda_i'$ of $\vec v_i$
is given by
\begin{equation}\label{linear eigenvalue perturbation} \lambda_i' =
\lambda_i + \epsilon\langle \vec u_j,T_1\vec u_i\rangle + \sheaf
O(\epsilon^2).
\end{equation} where here the braces $\langle\cdot,\cdot\rangle$
denote the usual inner product on $\bbr^n$.

To prove the above two equations, we can use a standard but important
linear algebra tool: the resolvent of a matrix.  Given any matrix $L$,
the resolvent $R(L;z)$ is a matrix-valued rational function on the
complex plane, defined by $R(L;z) = (Iz-L)^{-1}$.  One useful property
of the resolvent is that acts like a projection operator onto the
eigenspaces of $L$.  Specifically, if $\lambda$ is an eigenvalue of
$R(L;z)$ and $C$ is a closed contour in the complex plane, then for
\emph{any} constant vector $\vec v\in\bbr^n$, the integral
$$\frac{1}{2\pi i}\oint_C R(L;z)\vec v dz$$
is an eigenvector of $L$ with eigenvalue $\lambda$, as long as it is
nonzero.

Now since $T = T_0 + \epsilon T_1$, a simple calculation shows
\begin{align*} R(T;z) & = (I-R(T_0;z)T_1)^{-1}R(T_0;z)\\ & =
\sum_{m=0}^\infty \epsilon^m(R(T_0;z)T_1)^mR(T_0;z).
\end{align*} For each $i$, let $\lambda_i'$ be an eigenvalue of $T$
closest to $\lambda_i$.  Take $C_i$ to be a closed contour in the
complex plane containing $\lambda_i$ and $\lambda_i'$, but no other
(distinct) eigenvalue of $T$ or $T_0$.  Then we know that
$$\vec v_i := \frac{1}{2\pi i}\oint_{C_i} R(T,z)\vec u_idz$$
will be an eigenvector with eigenvalue $\lambda_i'$.  However, using
our expression for $R(T,z)$ we see that
\begin{align*} R(T,z)\vec u_i & = \sum_{m=0}^\infty
\epsilon^m(R(T_0;z)T_1)^m \frac{1}{z-\lambda_i}\vec u_i\\ & =
\frac{1}{z-\lambda_i}\left(\vec u_i + \sum_{j=i}^n
\frac{\epsilon\langle \vec u_j,T\vec u_i\rangle}{(z-\lambda_j)}\vec
u_j\right) + \sheaf O(\epsilon^2).
\end{align*} Therefore by Cauchy's residue theorem
$$\vec v_i = \vec u_i + \sum_{j:\lambda_j\neq\lambda i}\frac{\epsilon\langle \vec u_j,T_1\vec u_i\rangle}{\lambda_j-\lambda_i}\vec u_i + \sheaf O(\epsilon^2).$$
This proves Equation \ref{linear eigenvector perturbation}.  To get
the associated eigenvalue $\lambda_i'$, we can use the fact that
$$\lambda_i'\langle \vec u_i,\vec v_i\rangle = \langle \vec u_i,T\vec v_i\rangle.$$
From our expression for $\vec v_i$, we have $\langle \vec u_i,\vec
v_i\rangle = 1 + \sheaf O(\epsilon^2)$ and $\langle \vec u_i,T\vec
v_i\rangle = \lambda_i + \epsilon\langle\vec u_i,T_1\vec u_i\rangle +
\sheaf O(\epsilon^2)$.  Thus
$$\lambda_i'  = \lambda_i + \epsilon\langle \vec u_j,T_1\vec u_i\rangle + \sheaf O(\epsilon^2).$$
This proves Equation \ref{linear eigenvalue perturbation}.

\subsection*{Mathematical Justification for SP-MGM}
Let $G$ be a graph with $n$ vertices and let $C_1,\dots, C_k$ be the
natural subclusters of $G$.  To show mathematically how SP-MGM detects
the clusters $C_1,\dots, C_k$ from the graph $G$, we will consider two
subgraphs of $G$, which we will call $G_0$ and $G_1$, and which will
be formed by deleting the intercluster bonds and intracluster bonds of
$G$, respectively.  Also to simplify the notation, throughout this
section we will think about the Laplacian of $G$ as acting not on
$\bbr^n$, but rather on the space of real-valued functions on $G$.
For this reason, we will switch from talking about eigenvectors to
talking about eigenfunctions.  We will denote the Laplacians of $G$,
$G_0$ and $G_1$ as $L$, $L_0$ and $L_1$.  Clearly $L = L_0 + L_1$.

To better understand why SP-MGM should work well, we need to consider
in more detail the eigen-structure of the Laplacian $L_0$ of
$G_0$.  The eigenfunctions of $G_0$ clearly restrict to each cluster
$C_i$ to an eigenfunction on $C_i$.  Each cluster $C_i$ is connected,
and therefore has a kernel spanned by its indicator function.  Thus an
orthonormal basis for the kernel of $L_0$ is given by
$$f_i(x) = \left\lbrace\begin{array}{cc}n_i^{-1/2}, & x\in C_i\\ 0 & x\notin C_i\end{array}\right.$$
for $i=1,\dots,k$.  More generally, let $\lambda_{i,1}\leq\dots\leq
\lambda_{i,n_i}$ be the eigenvalues of $C_i$.  Then the $\lambda_{ij}$
are also eigenvalues of $L_0$ and there exists an orthonormal
eigenbasis for $L_0$ such that $f_{i,j}(x)$ has eigenvalue
$\lambda_{i,j}$ and is supported on $C_i$.  By definition, $f_{i,1}(x)
= f_i(x)$.

Assuming that the clusters are really, honestly clusters and not just
part of some artificial partition of $G$, we should expect the
spectral radius $\rho_1$ of $L_1$ to be small compared
to the spectral radius $\rho_0$ of $L_0$.  Then the eigendata of $L$
and $L_0$ will be related via the eigenvector perturbation theory
discussed above.  Each of the $f_i(x)$ will correspond to an
eigenfunction $f_i'(x)$ of the matrix $L$ via this perturbation
theory, and the associated eigenvalues should be expected to be the
smallest $k$ eigenvalues of the matrix $L$.  Moreover, this eigenvalue
perturbation theory gives us an estimate of $f_i(x)'$ relative to
$f_i(x)$.  If we can show that
$$\|f_i(x)-f_i'(x)\|_\infty < \frac{1}{4n_i^{1/2}},$$
then the location of the maximum gap in $f_i(x)$ and the location of
the maximum gap in $f_i(x)$ must be the same!  In this case, SP-MGM is
guaranteed to identify a partition which correctly separates one of
the clusters from the others.

From our eigenvalue perturbation theory, we know that for
$m=1,\dots,k$
$$f_m'(x) = f_m(x) + \sum_{i=1}^k\sum_{j=2}^{n_i} \frac{\langle f_{i,j}(x),L_1 f_{m1}(x)\rangle}{\lambda_{i,j}}f_{i,j}(x)$$
to order $\sheaf O(\epsilon^2)$, where $\epsilon = \rho_1/\rho_0$.
Since $f_{ij}$ is supported on $C_i$ and $L_1$ sends $f_m$ to a
function supported on the complement of $C_m$, we find that to order
$\sheaf(\epsilon^2)$
$$\|f_m'(x) - f_m(x)\| \leq \max_{i\neq m}(n_i-1)\frac{\rho_1}{\lambda_{i,2}}.$$
Thus we arrive at the conclusion that if $\epsilon = \rho_1/\rho_0$ is
small and for all $m$ we have
$$\max_{i\neq m}(n_i-1)\frac{\rho_1}{\lambda_{i,2}} < \frac{1}{4n_m^{1/2}}.$$
Then SP-MGM will correctly identify the desired clusters.

\bibliographystyle{plain} \bibliography{cluster}

\begin{IEEEbiography}[{\includegraphics[width=1in,height=1.25in,clip,keepaspectratio]{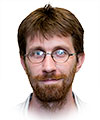}}]{W.R. Casper}
received BS degrees in mathematics and physics, as well as an MS in
mathematics from North Dakota State University in 2010.  He later
received a PhD in mathematics from the University of Washington,
Seattle in 2017.  He is currently a postdoctoral researcher in the
Department of Mathematics at Louisiana State University in Baton
Rouge.  His research interests are diverse, and include integrable
systems, algebraic geometry, noncommutative algebra, geophysical fluid
dynamics, computational physics and machine learning. 
\end{IEEEbiography}

\begin{IEEEbiography}[{\includegraphics[width=1in,height=1.25in,clip,keepaspectratio]{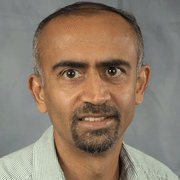}}]{Balu
    Nadiga} is a scientist at the Los Alamos National Lab. With a
  specialization in fluid dynamics, he has worked on problems ranging
  from climate and ocean circulation to turbulence modeling and
  multiphase flows. His other interests include dynamical systems,
  uncertainty quantification and Bayesian analysis, and using machine learning
  for turbulence modeling.
\end{IEEEbiography}

\end{document}